\documentclass[a4paper,english]{svmult}
\usepackage[utf8]{inputenc}
\usepackage{graphicx}
\usepackage{hyperref}
\usepackage{float, subfig}
\usepackage{listings}
\usepackage{amssymb, bm}
\usepackage{amsmath,amsfonts,amssymb}
\usepackage{algorithm}
\usepackage{lipsum}
\usepackage{algorithmic}
\usepackage{xcolor} 

\usepackage[inline]{enumitem}
\usepackage[english]{babel}
\usepackage{microtype}
\begin{document}

\title*{Co-clustering based exploratory\\ analysis of mixed-type data tables}

\author{Aichetou Bouchareb, 
        Marc Boullé,
        Fabrice Clérot and 
				Fabrice Rossi
        }

\institute{Aichetou Bouchareb, Marc Boullé and  Fabrice Clérot : \at Orange Labs, 2 Avenue Pierre Marzin 22300 Lannion - France, \email{firstname.lastname@orange.com} 
\and Fabrice Rossi, Aichetou Bouchareb : \at SAMM EA 4534 - University of Paris 1 Panthéon-Sorbonne, 90 rue Tolbiac 75013 Paris - France, \email{firstname.lastname@univ-paris1.fr}}
\maketitle

\abstract{Co-clustering is a class of unsupervised data analysis techniques
  that extract the existing underlying dependency structure between the
  instances and variables of a data table as homogeneous blocks. Most of those
  techniques are limited to variables of the same type. In this paper, we
  propose a mixed data co-clustering method based on a two-step
  methodology. In the first step, all the variables are binarized according to
  a number of bins chosen by the analyst, by equal frequency discretization in
  the numerical case, or keeping the most frequent values in the categorical
  case. The second step applies a co-clustering to the instances and the
  binary variables, leading to groups of instances and groups of variable
  parts. We apply this methodology on several data sets and compare with the
  results of a Multiple Correspondence Analysis applied to the same data.}

\section{Introduction}
Data analysis techniques can be divided into two main categories: supervised
analysis, where the goal is to predict a mapping between a set of input
variables and a target output variable, and unsupervised analysis where the
objective is to describe the set of all variables by  uncovering the
underlying structure of the data. This is often achieved by identifying dense
and homogeneous clusters of instances, using a family of techniques called {\it clustering}.

{\it Co-clustering} (\cite{good1965, hartigan1975}), also called
cross-classification, is an extension of the standard clustering approach. It
is a class of unsupervised data analysis techniques that aim at simultaneously
clustering the set of instances and the set of variables of a data table.

Over the past years, numerous co-clustering methods have been proposed (for
example, \cite{bock1979}, \cite{govaert1983}, \cite{dhillon2003}, and
\cite{govaert2013}). These methods differ on several axes including: data
types, clustering assumptions, clustering techniques, expected results,
etc. In particular, two main families of methods have been extensively
studied: matrix reconstruction based methods where the co-clustering is viewed
as a matrix approximation problem, and mixture model based methods where the
co-clusters are defined by latent variables that need to be estimated (for a
full review of co-clustering techniques, readers are referred to
\cite{brault2015b}). The typical models used in mixture based approaches are
Gaussian for numerical data, multinomial for categorical data and Bernoulli
for binary data.

\begin{figure}[htbp!]
    \centering
  \includegraphics[width=0.2\textwidth]{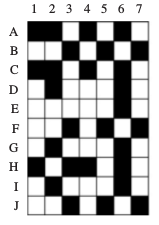}
	\qquad
  \includegraphics[width=0.15\textwidth]{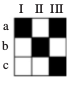}
    \caption{An illustration of a co-clustering where the original binary data table is on the left and the co-clustered binary table is on the right.}
    \label{fig1_Coclust_Example}
\end{figure}

Figure~\ref{fig1_Coclust_Example} shows an example of a binary data table representing $n=10$
instances and $m=7$ variables  (\cite{govaert2008}) and the binary table of
co-clusters resulting from a co-clustering into $3\times 3=9$ co-clusters. The
table of co-clusters provides a summary of the original data and allows
to view the main associations between the set of instances and the set of
variables.

Co-clustering methods are naturally limited to homogeneous data where all
variables are of the same nature: binary, numerical or categorical.  In the
present paper, we propose to extend these exploratory analysis methods to
the case of mixed-type data using a two-step methodology.  The first step
consists in binarizing the data using a number of parts, given by the analyst,
using equal frequency discretization in the case of numerical variables and
keeping the most frequent values in the case of categorical variables. The
second step consists of using a co-clustering method between the instances and
variable parts, leading to a partition of the instances on one hand and a
partition of the variable parts on the other hand.

Given a number of parts, our objective is to require no further parameters such as the number of instance clusters and the number of variable part clusters. Therefore, in the  co-clustering step, 
we use the MODL approach  (\cite{Boulle2011}) for its non parametric nature,
its efficiency for extracting correlation structures from the data, its
scalability and its robustness to overfitting induced by the embedded
regularization.

Since we are in the context of exploratory analysis of a mixed-type data
table, we compared our methodology to the most widely used factor analysis
method in case of the presence of categorical variables: Multiple
Correspondence Analysis (MCA). Indeed, MCA is factor analysis technique that
enables one to extract and analyze the correlations between categorical variables
while performing a typology of instances. It enables the instances and the
variables to be handled in a complementary manner by duality where groups of
instances can be interpreted using variables and vice-versa. These aims of MCA
are thus consistent with the goals of co-clustering, hence the usefulness of
such comparison.

The remainder of this paper is organized as follows. In section~\ref{secMODL}
we give an outline of the MODL approach for co-clustering, then in
section~\ref{approche} we illustrate our proposed methodology for
co-clustering mixed-type data tables. In section~\ref{acm}, we present a
summary of the MCA basics.  Section~\ref{experimentations} presents the
experimental results along with a comparative analysis. Finally, conclusions
and future work are presented in section~\ref{conclusion}.

\section{MODL based co-clustering of two categorical variables}
\label{secMODL}
This section presents a summary of the MODL  approach
  (\cite{Boulle2011}) that clusters simultaneously the \emph{values} of two
  categorical variables $X$ and $Y$. 
In definition~\ref{JointDensityModel}, we introduce a family of models for
estimating the joint density of two categorical variables, based on
partitioning the values of each variable into groups of values (hence MODL
performs value oriented co-clustering). We then present the evaluation
criterion for these models in theorem~\ref{JointDensityTheorem}.

\begin{definition}
\label{JointDensityModel}
A co-clustering model of two categorical variables is defined by:
\begin{itemize}
	\item a number of groups for each variable,
	\item the partition of the values of each variable into groups of values,
	\item the distribution of the instances of the data over the cells of the
resulting data grid,
	\item for each variable and each group, the distribution of the instances of the
group on the values of the group.
\end{itemize}
\end{definition}

\paragraph{Notations:}
\begin{itemize}
	\item $N$: number of instances
	\item $V, W$: number of values for each variable (assumed known)
	\item $I, J$: number of groups for each variable (unknown)
	\item $G=I J$: number of cells in the resulting data grid
	\item $m_{i.}, m_{.j}$: number of values in group $i$ (resp. $j$)
	\item $n_{v.}, n_{.w}$: number of instances for value $v$ (resp. $w$)
	\item $n_{vw}$: number of instances for the pair of values $(v, w)$
	\item $N_{i.}, N_{.j}$: number of instances in the group $i$ (resp. $j$)
	\item $N_{ij}$: number of instances in the cell $(i, j)$ of the data grid
\end{itemize}

Every model from the set of models in definition~\ref{JointDensityModel} is
completely defined by the choice of $I$, $J$, $N_{ij}$, $n_{v.}$, $n_{.w}$,
and the partition of the values of each variable to groups (clusters). In the
co-clustering context, these parameters correspond to the number of clusters
per variable,  the multinomial distribution of the instances on the
co-clusters, and the parameters of the multinomial distributions of the
instances of each variable cluster on the values of the cluster. Notice that these
parameters are optimized by the algorithm and not fixed the analyst: by using
MODL we will not add any additional user chosen parameter to the data pre-processing parameter. 

The number of values in each cluster $m_{i.}$ and $m_{.j}$ result from the partition of the values of each variable into the defined number of clusters. Similarly, the number of instances per cluster $N_{i.}$ and $N_{.j}$ are derived by summation from the number of instances per co-cluster  ($N_{i.}= \sum_j{N_{ij}}$ and $N_{.j}= \sum_i{N_{ij}}$). 

In order to select the best model, a MAP based criterion is chosen: we
maximize the probability of the model given the data
$P(M|D)=\frac{P(M)P(D|M)}{P(D)}$. We use a prior distribution on the model
parameters that exploits the natural hierarchical nature of the
parameters. The distribution is uniform at each level of the hierarchy. In
practice, it serves as a regularization term which prevents the optimization
from selecting systematically a high number of groups, for instance. 

 Using the formal definition of the joint density estimation models and its prior hierarchical distribution, the Bayes formula enables us to compute the exact probability of a model given the data, which leads to theorem \ref{JointDensityTheorem}.

\begin{theorem} 
\label{JointDensityTheorem}
Among the set of models, a co-clustering model distributed according to a uniform hierarchical prior is Bayes optimal if its evaluation according to the following criteria is minimal (\cite{Boulle2011}):
{\scriptsize{
\begin{equation}
\label{JointDensityCriterion}
\begin{split}
c(M) = & \log V + \log W + \log B(V, I) + \log B(W, J)\\
 +& \log \binom {N + G - 1} {G - 1}
  + \sum_{i=1}^{I} {\log \binom {N_{i.} + m_{i.} - 1} {N_{i.} - 1}}
  + \sum_{j=1}^{J} {\log \binom {N_{.j} + m_{.j} - 1} {N_{.j} - 1}}\\
+ & \log N! - \sum_{i = 1}^{I} {\sum_{j = 1}^{J} {\log N_{ij}!}}
 + \sum_{i = 1}^{I} {\log N_{i.}!} 
  + \sum_{j = 1}^{J} {\log N_{.j}!}
  - \sum_{v = 1}^{V} {\log n_{v.}!}
  - \sum_{w = 1}^{W} {\log n_{.w}!}
\end{split}
\end{equation}
}}
where $B(V, I)$ is the number of ways of partitioning a set of $V$ elements into $I$ nonempty groups which can be written as a sum of the Stirling number of the second kind: $B(V, I)=\sum\limits_{i=1}^I{S(V,i)}$.
\end{theorem}

The first  line of this criterion corresponds to the prior distribution of choosing the numbers of groups and to the partition of the values of each variable to the chosen number of groups. The second line represents the
specification of the parameters of the multinomial distribution of the $N$ instances on the  $G$ cells of the data grid and the specification of the multinomial distribution of the instances of each group on the values of the group. The third line corresponds to the likelihood
of the distribution of the instances on the data grid cells and the likelihood of the distribution of the instances per group over the values in the group,  by the mean of a multinomial term.

The estimation of the joint density of two categorical variables distributed
according to hierarchical parameter priors is implemented in the software
Khiops\footnote{The Khiops tool is available as a shareware at
  \url{www.khiops.com/}.}. We use this software for our experiments presented
in section~\ref{experimentations}. The detailed formulation of the
  approach as well as optimization algorithms and asymptotic properties can be
  found in \cite{Boulle2011}.

\section{Mixed-type data co-clustering}
\label{approche}
In this section we present our two-step approach. The first step is described in
Sections \ref{pretraitement} and \ref{sec:data-transformation} and consists in  binarizing the numerical and
categorical variables. The second step leverages the MODL approach to perform
a co-clustering of the instances $\times$ binarized variables data, see
Section \ref{sec:co-clustering-co}. 

\subsection{Data pre-processing}
\label{pretraitement}
The first step of our methodology consists of binarizing all
variables using a user parameter $k$, which represents the maximal number of
parts per variable. In the case of a numerical variable, these parts are the
result of an unsupervised discretization of the range of the variable into $k$
intervals with equal frequencies. In the case of a categorical variable, the
$k-1$ most frequent values define the first $k-1$ parts while the $k^{th}$
part receives all the other values.  An alternative discretization
  is with equal bins. However, frequency based discretization reinforces the
  robustness of the approach and minimizes the effect of outliers  if present
  in the data  (both outlier instances and variable values).

The parameter $k$ defines the maximal granularity at which the
  analysis can be performed. A good choice of $k$ is related to a trade off
  between the fineness of the analysis, the time required to compute the
  co-clustering of the second step, and the interpretability of the
  co-clustering results.  The computational cost of the MODL
    co-clustering in the worst case is in $\mathcal{O}(N\sqrt{N}\log N)$ where $N$ is the total
    number of instances (in our case, $N=n\times m$, see Section
    \ref{sec:data-transformation}), but the observed computation time tends to decrease with smaller $k$, when data is far from the worse case.  Also the size of the
  data set and its complexity can be taken as an indicator, small values are
  probably sufficient for small and simple data sets while for larger ones, it
  would be wise to choose a larger parameter $k$. Nevertheless, we recommend
  to start with high values of $k$ since it gives a detailed description of
  the data. Starting from a detailed description, the MODL approach will group
  the variable parts that needed not to be separated in the same cluster which
  can only enhance the level of correspondence of the resulting co-clustering
  to the original data, without much loss of information.
 
  One should note, however, that the granularity parameter $k$ is far less
  restrictive than other common parameters such as the number of instance
  clusters and the number of variable clusters, commonly used in the vast
  majority of co-clustering methods. In our experiments, we used $k=5$ for a
  small data set and $k=10$ for a relatively large one.

If we take the Iris database for example, the output of the binarization step, for $k=5$, is illustrated in table~\ref{iris_k5}.
\begin{table}[!htb]
\centering
\begin{tabular}{ccccc}
\hline
SepalLength  & SepalWidth & PetalLength & PetalWidth & Class\\
\hline
$]-\infty;5.05]$ & $]-\infty;2.75]$ & $]-\infty;1.55]$ & $]-\infty;0.25]$& Iris-setosa \\
$]5.05;5.65]$ & $]2.75;3.05]$ & $]1.55;3.95]$ &  $]0.25;1.15]$& Iris-versicolor \\
$]5.65;6.15]$ & $]3.05;3.15]$ & $]3.95;4.65]$& $]1.15;1.55]$& Iris-virginica \\
$]6.15;6.55]$ & $]3.15;3.45]$ & $]4.65;5.35]$ & $]1.55;1.95]$&\\
$]6.55;+\infty[$ & $]3.45;+\infty[$ & $]5.35;+\infty[$ & $]1.95;+\infty[$&\\
\hline
\end{tabular}
\caption{The output of the discretization step for $k=5$}\label{iris_k5}
\end{table}

\subsection{Data transformation}\label{sec:data-transformation}
The MODL approach (\cite{Boulle2011}), summarized in Section~\ref{secMODL},
has been chosen because it is non parametric, effective, efficient, and
scalable. Although designed for joint density estimation, MODL has also been
applied to the case of instances$\times$binary-variables. An example of such
application is that of a large corpus of documents, where each document is
characterized by tens of thousands of binary variables representing the usage
of words. In this case, the corpus of documents is transformed beforehand into
a representation in the form of two variables \emph{IdText} and
 \emph{IdWord}.

\begin{table}[h]
\begin{center}
\begin{tabular}{cl}
\hline
\emph{IdInstance} & \emph{IdVarPart}\\
\hline
$I1$ & \emph{SepalLength$]5.05;5.65]$} \\
$I1$ & \emph{SepalWidth$ ]3.45;+\infty[$} \\
$I1$ & \emph{PetalLength$]-\infty;1.55]$} \\
$I1$ & \emph{PetalWidth$ ]-\infty;0.25]$} \\
$I1$ & \emph{Class\{Iris-setosa\}} \\
$I2$ & \emph{SepalLength $]-\infty;5.05]$} \\
$I2$ & \emph{SepalWidth$]2.75;3.05]$} \\
$I2$ & \emph{PetalLength$]-\infty;1.55]$} \\
$I2$ & \emph{PetalWidth$]-\infty;0.25]$} \\
$I2$ & \emph{Class\{Iris-setosa\}} \\
\hline
\end{tabular}
\caption{The first 10 instances of the binarized Iris database}
\label{wrapped_fig1}
\end{center}
\end{table}

In the same manner, we transform the binarized database into two variables
\emph{IdInstance} and \emph{IdVarPart} by creating, for each instance, a
record per variable that logs the link between the instance and its variable
part. The set of $n$ initial instances characterized by $m$ variables is thus
transformed into a new data set of $N=n\times m$ instances and two categorical
variables, the first of which contains $V=n$ values and the second containing,
at most, $W=m \times k$ values.  For instance, in the Iris database, this
transformation results in two columns of $750$
instances. Table~\ref{wrapped_fig1} shows the first ten instances. Notice that
after the transformation, the algorithm cannot leverage two aspects of the
data: the actual value taken by a variable inside a variable part and the
original links between variable parts. In other words, the fact that
\emph{SepalLength$]5.05;5.65]$} and \emph{SepalLength $]-\infty;5.05]$} both
refer to the same original variable is not leveraged by MODL. 
  
\subsection{Co-clustering and co-cluster interpretation}\label{sec:co-clustering-co}
Now that our data is represented in the form of two categorical variables, we
can apply MODL to find a model estimating the joint density between these two
variables.  This results in two partitions of the values of the
  newly introduced categorical variables. Clusters of values of
  \emph{IdInstance} are in fact clusters of instances while clusters of values
  of \emph{IdVarPart} are clusters of variable parts. Thus the results is a
  form of co-clustering in which variables are clustered at the level of parts
  rather than globally.  In the resulting co-clustering, the instances of the
original database (values of the variable \emph{IdInstance}) are grouped if
they are distributed similarly over the groups of variables parts (values of
the variable \emph{IdVarPart}), and vice-versa.

 When the optimal co-clustering is too detailed, coarsening of the
  partitions can be implemented by merging clusters (of objects or variable
  parts) in order to obtain a simplified structure.  While this model
  coarsening approach can degrade the co-clustering quality, the induced
  simplification enables the analyst to gain insight on complex data at a
  coarser level, in a way similar to exploration strategies based on
  hierarchical clustering. The dimension on which the merging is performed and
  the best merging are chosen optimally at each coarsening step with regards
  to the minimum divergence from the optimal co-clustering, measured by the
  difference between the optimal value of the criterion and the valued
  obtained after merging to clusters. 

\section{Multiple Correspondence Analysis}
\label{acm}
Factor analysis is a set of statistical methods, the purpose of which is to
analyze the relationships or associations that exist in a data table, where
rows represent instances and columns represent variables (of any type).

The main purpose of factor analysis is to determine the level of similarity
(or dissimilarity) between groups of instances (problem classically treated by
clustering) and the level of associations (correlations) between the observed
variables. Multiple correspondence analysis is a factor analysis technique
that enables one to  analyze the correlations between multiple categorical variables
while performing a typology (grouping) of instances and variables in a
complementary manner.  

\subsection{MCA in practice}
Let $\mathbf{x}=(x_{ij}, i\in I, j\in J)$ be the instance$\times$variables
data table, where $I$ is the set of $n$ studied objects and $J$ is the set of
$p$ categorical variables (with $m_j$ categories each) characterizing the
objects. Since mathematical operations would not make sense in categorical
variables, MCA uses an  indicator matrix called complete disjunctive table
(CDT) which is a juxtaposition of $p$ indicator matrices of all variables
where rows represent the instances and columns represent the categories of the
variable. This CDT can be considered as a contingency table between instances
and the set of all categories in the data table.  

For a given CDT, $T$, the sum of all elements of each row is equal to the number $p$ of variables, the sum of all elements of a column $s$ is equal to the marginal frequency $n_s$ of the corresponding category, the sum of all columns in each indicator matrix is equal to 1, the sum of all elements in $T$ is equal to $np$, the matrix of row weights is given by $r = \frac{1}{n}I$, and the column weights are given by the diagonal matrix $D =  diag(D_1,D_2, \ldots, D_p)$ where each $D_j$ is the diagonal matrix containing the marginal frequencies of all categories of the $j^{th}$ variable.

\subsection{Main mathematical results for MCA}
The principal coordinates of categories are given by the eigenvectors of $\,\frac{1}{p} \mathbf{D}^{-1}\mathbf{T}^t\mathbf{T}$, which are the solutions of the equation:   
$$\frac{1}{p}\mathbf{D}^{-1}\mathbf{T}^t\mathbf{T} \mathbf{a} =\mu \mathbf{a}$$

The principal coordinates of instances are given by the eigenvectors of $\,\frac{1}{p}\mathbf{T} \mathbf{D}^{-1} \mathbf{T}^t$, which are the solutions of the equation:  
 $$\frac{1}{p} \mathbf{T} \mathbf{D}^{-1} \mathbf{T}^t \mathbf{z}=\mu \mathbf{z}$$

We deduce (\cite{saportabook}) the transition formulas given by 
$z = \frac{1}{\sqrt{\mu}} \frac{1}{p} \mathbf{T}\mathbf{a}$ and $\mathbf{a} = \frac{1}{\sqrt{\mu}} \mathbf{D}^{-1} \mathbf{T}^t\mathbf{z}$, which describes how to pass between the coordinates.

\paragraph{Note that:}
 \begin{itemize}
 \item the total inertia is equal to ($\frac{m}{p} - 1$), where $m$ is the total number of categories.
 \item the inertia of all the  $m_j$ categories in the $j^{th}$ variable is equal to $\frac{1}{p}(m_j-1)$. Since the contribution of a variable to the total inertia is proportional to the number of categories in the variable, it is preferable to require all variables to have the same number of categories, hence the utility of the pre-processing step (section \ref{pretraitement}).

\item the contributions of an instance $i$ and of a category $s$ to a principal axis are given by:
  $$Ctr_h(i) =\frac{1}{n} \frac{z_{ih}^2}{\mu_h} \text{ et } Ctr_h(s) =\frac{n_s}{np} \frac{a_{ih}^2}{\mu_h}$$
  
  \item the contribution of a variable to the inertia of a factor is equal to the sum of contributions of all categories  in the variable to that same axis. This  contribution measures the level of correlation between the variable and the principal axis.
\end{itemize}

MCA can be used to simultaneously analyze categorical and numerical variable. To do so, we follow the classic approach of decomposing the range of each numerical variable into intervals.

\section{Experiments}
\label{experimentations}
We start the experiments by comparing our  methodology (Section~\ref{approche}) with MCA (Section~\ref{acm}) using the Iris database for didactic reasons, then we evaluate our approach using the Adult database (\cite{datasets}) to evaluate its scalability.

\subsection{The case study:  Iris database}
The Iris database consists of $n=150$ instances and $m=5$ variables, four numerical and one categorical.

\subsubsection{Co-clustering}

\begin{figure}[htbp!]
\centering
  \includegraphics[width=0.65\textwidth]{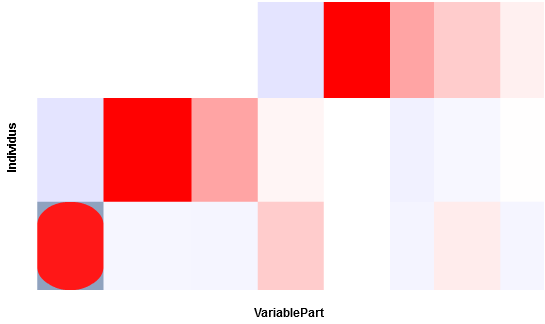}
  \caption{Coclustering pour la base Iris}
\label{figIrisCC}
\end{figure}

After binarizing the Iris data using a granularity of $k=5$ parts and applying
the MODL co-clustering method, we found that the optimal grid consists of $3$
clusters of instances and $8$ clusters of variable
parts. Figure~\ref{figIrisCC} illustrates this grid where rows represent the
instance clusters and columns represent the variable part clusters. The mutual
information between the two dimensions can be visualized in each cell, where
the red color represents an over-representation of the instances compared to
the case where the two dimensions are independent and the blue color
represents an under representation.    

The three instance clusters, shown in figure~\ref{figIrisCC}, can be
characterized by the types of flowers of which they are composed and by the
most represented variable parts per cluster (the red cell of each row of the
grid):

\begin{itemize}
\item in the first row: a cluster of 50 flowers, all of the class \emph{Iris-setosa} and characterized by the variable parts:  \emph{Class\{Iris-setosa\}}, \emph{PetalLength$]-inf;1.55]$} and \emph{PetalWidth$]-inf;0.25]$},

	\item in the second row: a cluster of 54 flowers,  50 of which are of the class \emph{Iris-virginica}, and  characterized by the following variable parts: \emph{Class\{Iris-virginica\}}, \emph{PetalLength$]5.35;+inf[$}, \emph{PetalWidth\-$]1.95;+inf[$} and \emph{PetalWidth$]1.55;1.95]$},

	\item the third row: a cluster of 46 flowers, all of the class
          \emph{Iris-versicolor}, and caracterized by the variable parts:
          
          \emph{ Class\{Iris-versicolor\}}, \emph{ PetalLength$]3.95;4.65]$} and
          \emph{ PetalWidth$]1.15;1.55]$}.
\end{itemize}
Notice first that, as expected, the methodology enables us to group variable (parts)
of different nature in the same cluster. 

The three instance clusters are easily understandable as they represent the
\emph{small}, \emph{large} and \emph{medium} flowers respectively. These
clusters are mainly explained by three clusters of variable parts containing
the variables \emph{Class}, \emph{PetalLength} and \emph{PetalWidth}. In fact
it is well known that in the Iris data set, the three classes are well
separated by the Petal variables. This is reflected here by the grouping of
the variables as well as by the instance clusters. 

Conversely, looking at the clusters of variable parts, one can distinguish two
non informative clusters (the fourth and eighth columns which are the two
columns with the least contrast), which are based essentially on the variable
\emph{SepalWidth}:
\begin{itemize}
\item the fourth column  contains the parts:

 \emph{SepalWidth$]-\infty;2.75]$}, \emph{SepalWidth$]2.75;3.05]$}, and \emph{SepalLength$]5.65;6.15]$},

\item the eighth column contains the parts:

 \emph{SepalWidth$]3.05;3.15]$} and  \emph{SepalWidth$]3.15;3.45]$}.
\end{itemize}
The small values of \emph{SepalWidth} (fourth column) are slightly
over-represented by the cluster of instances associated to the classes
\emph{Iris-versicolor} and \emph{Iris-virginica} while the intermediate values
(eighth column) are slightly over-represented for the cluster of instances
associated to \emph{Iris-versicolor}.

\subsubsection{MCA analysis}
MCA analysis is performed based on the same  data binarization as previously.  

\begin{figure}[htbp!]
\centering
  \includegraphics[width=0.45\textwidth]{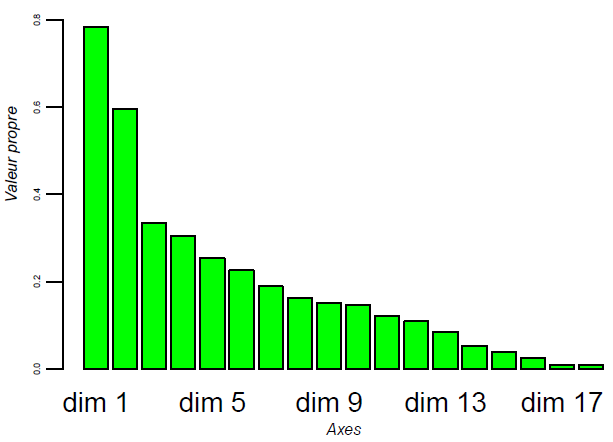}
  \includegraphics[width=0.45\textwidth]{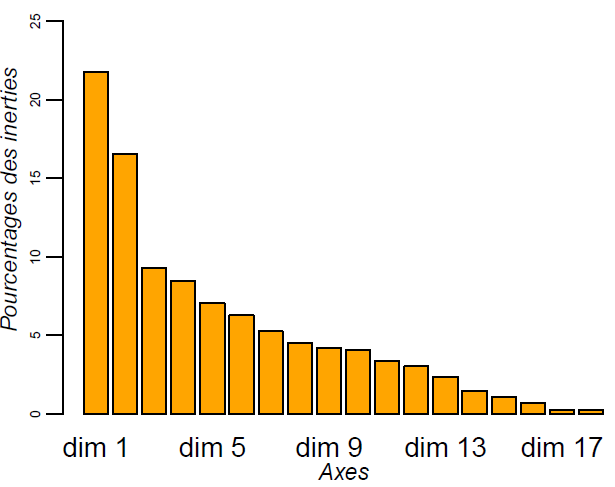}
  \caption{Histogram of eigenvalues (on the left) and the percentage of variance captured by the axes in the MCA analysis of Iris}
\label{figIrisACMHisto}
\end{figure}

The distribution of eigenvalues (Figure~\ref{figIrisACMHisto}) indicates that the first two principal axes do capture enough information with a cumulative variance of 38.30\%. Therefore, we will limit our analysis to the first factorial plan. 

\begin{figure}[htbp!]
\centering
  \includegraphics[width=0.49\textwidth]{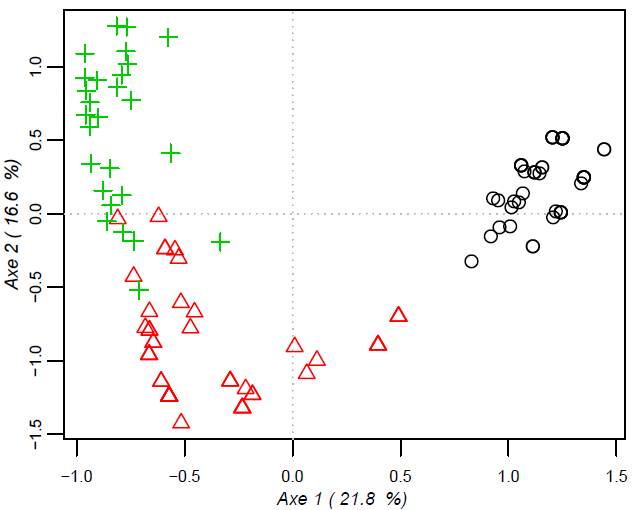}
  \includegraphics[width=0.49\textwidth]{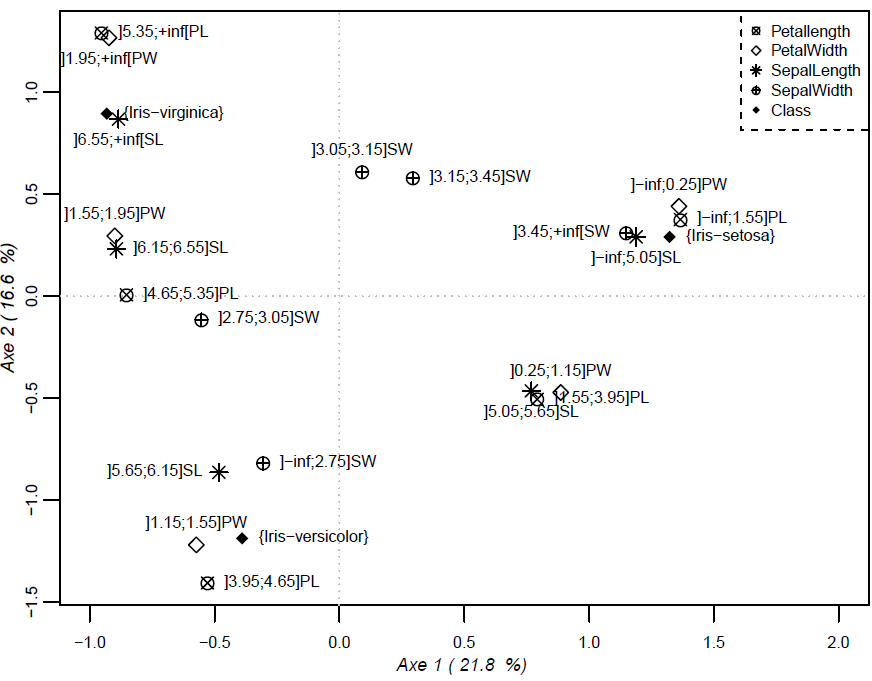}
  \caption{Projection of the set of instances and variable parts on the first factorial plan}
\label{figIrisACM}
\end{figure}

The comparison between the projection of variables  (figure~\ref{figIrisACM} on the right) and the projection of instances (figure~\ref{figIrisACM} on the left), over the first factorial plan, reveals some clear correlations:
\begin{itemize}
\item in the top left, \emph{Iris-virginica} is correlated with high values of \emph{PetalLength} (greater than 4.65), high values of  \emph{PetalWidth} (greater than 1.55) and high values of \emph{SepalLength} (greater than 6.15),

\item  on the right, \emph{Iris-setosa} is strongly correlated with low values of \emph{PetalLength} (less than 3.95), low values of \emph{PetalWidth} (less than 1.15) and low values of \emph{SepalLength} (less than 5.05),

\item in the bottom left,  \emph{Iris-versicolor} is correlated with intermediate values of  \emph{PetalLength}, \emph{PetalWidth} and \emph{SepalWidth}.
\end{itemize}

The projection of instances (on the left of figure~\ref{figIrisACM}) shows a mixture between \emph{Iris-virginica} and \emph{Iris-versicolor}.
These results are identical to those found using the co-clustering analysis. 

The variable parts issued from \emph{SepalWidth} are weakly correlated with the others and contribute less the first factorial plan: the small values (less than 3.05) are associated with the mixture zone between  \emph{Iris-virginica} and \emph{Iris-versicolor}, the intermediate values  (between 3.05 and 3.45) have their projections in between \emph{Iris-virginica} and \emph{Iris-setosa} (they are therefore present in both flowers). 

These results are also in agreement with the results deduced from the co-clustering (see the above interpretation of the fourth and eighth columns in the co-clustering).

Finally, on this didactic example where the results of MCA are easily interpretable, a good agreement emerges between the MCA and the proposed co-clustering approach.  

\subsection{The case study:  Adult database}
The Adult database is composed of $n=48842$ instances represented by  $m=15$ variables, 6 numerical and 9 categorical. 
\subsubsection{Co-clustering}\label{co-clustering_adult}
\begin{figure}[htbp!]
\centering
  \includegraphics[width=0.45\textwidth]{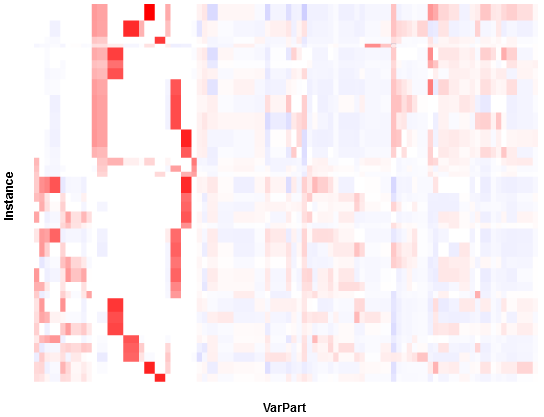}
	\quad
  \includegraphics[width=0.45\textwidth]{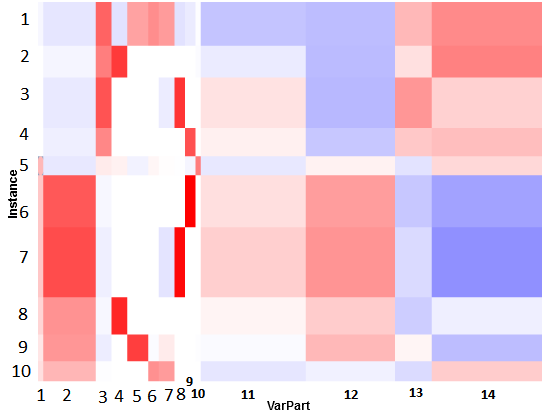}
  \caption{Co-clustering of the Adult database, with 100\% of information (on the left) and 70\% of  information (on the right).}
\label{figAdultCC}
\end{figure}

When the Adult data is binarized, using $k=10$, and the transformation into
two variables is performed as presented in Section~\ref{approche}, we obtain a
data set of $N\approx 750,000$ rows and two columns: the \emph{IdInstance}
variable containing around $n\approx 50,000$ values (corresponding to the
initial instances) and the \emph{IdVarPart} variable containing
$m \times k \approx 150$ values (corresponding to the variable parts). The
co-clustering algorithm is an \emph{anytime}, regularly issuing its quality
index (the achieved level of compression). For the Adult database, the
co-clustering takes about 4 mn for a first quality result (a time beyond which
the level of compression does not improve significantly). However, we proceed
with the optimization for about an hour which results in around $5\%$ of
improvement in the log-likelihood of the model. The obtained result is very
detailed, with 34 clusters of instances and 62 clusters of variable parts. In
an exploratory analysis context, this level of detail hinders the
interpretability. In our case, the results can be simplified by iteratively
merging the rows and columns of the finest clusters until reaching a
reasonable percentage of the initial amount of
information. Figure~\ref{figAdultCC} presents the co-clustering results with
$34\times 62$ clusters (on the left), which represents $100\%$ of the initial
information, and a simplified version with $ 10\times 14$ clusters preserving
$70\%$ of the initial information in the data.

The first level of retrieved patterns appears clearly when we consider dividing the clusters of instances  into two parts, visible on the top half and the bottom half of the co-clustering cells presented in figure~\ref{figAdultCC}. The instance clusters in the top half are mainly men with a good salary, with an over-representation of the variable part clusters containing \emph{sex\-\{Male\}}, \emph{relationship\-\{Husband\}}, \emph{relationship\-\{Married...\}}, \emph{class\-\{More\}}, \emph{age\-$]45.5;51.5]$}, \emph{age\-$]51.5;58.5]$},
\emph{hoursPerWeek\-$]48.5;55.5]$}, \emph{hoursPerWeek\-$]55.5;+\infty[$}. The instance clusters in the bottom half are mainly for women or rather poor unmarried men, with an over-representation of the variable part clusters containing \emph{class\-\{Less\}}, \emph{sex\-\{Female\}}, \emph{maritalStatus\-\{Never-married\}}, \emph{maritalStatus\-\{Divorced\}}, \emph{relationship\-\{Own-child\}}, \emph{relationship\-\{Not-in-family\}}, \emph{relationship\-\{Unmarried\}}.

In the left side figure, the instance cluster with the most contrast (hence the most informative) is on the first row and it can easily be interpreted by the over-represented variable part clusters in the same row:  
\begin{itemize}
	\item 
	\emph{relationship\{Husband\}}, \emph{relationship\{Married...\}},

	\item 
\emph{educationNum$]13.5;+\infty[$},
\emph{education\{Masters\}},

	\item 
\emph{education\{Prof-school\}},

	\item 
\emph{sex\{Male\}},

	\item 
\emph{class\{more\}},

	\item 
\emph{occupation\{Prof-specialty\}},

	\item 
\emph{age$]45.5;51.5]$},
\emph{age$]51.5;58.5]$},

	\item 
\emph{hoursPerWeek$]48.5;55.5]$},
\emph{hoursPerWeek$]55.5;+\infty[$}.
	
\end{itemize}

It is therefore a cluster of around 2000 instances, with mainly married men with rather long studies, working in the field of education, at the end of their careers, working extra-time with good salary.

In the right side figure, the most contrasted clusters of variable parts, hence the most informative, are those presented by the columns $4$ to $9$. These contain only variable parts issued from the variables \emph{education} and \emph{educationNum} which are the most structuring variables for this data set.

\begin{itemize}
	\item  
\emph{educationNum$]11.5;13.5]$},
\emph{education\{Assoc-acdm\}},
\emph{education\{Bachelors\}} (the $4^{th}$ column),

	\item 
\emph{educationNum$]-\infty;7.5]$},
\emph{education\{10th\}},
\emph{education\{11th\}},
\emph{education\{7th-8th\}}(the $5^{th}$ column),

	\item 
\emph{educationNum$]13.5;+\infty[$},
\emph{education\{Masters\}}(the $6^{th}$ column),

	\item 
\emph{educationNum$]10.5;11.5]$},
\emph{education\{Assoc-voc\}},
\emph{education\{Prof-school\}} (the $7^{th}$ column),

	\item 
\emph{educationNum$]7.5;9.5]$},
\emph{education\{HS-grad\}}(the $8^{th}$ column),

	\item  
\emph{educationNum$]9.5;10.5]$},
\emph{education\{Some-college\}}(the $9^{th}$ column).
\end{itemize}

The variables \emph{education} and  \emph{educationNum} are, respectively, categorical and numerical, very correlated as their variable part clusters seem particularly consistent.

\subsubsection{MCA analysis}
\begin{figure}[htbp!]
\centering
  \includegraphics[width=0.4\textwidth]{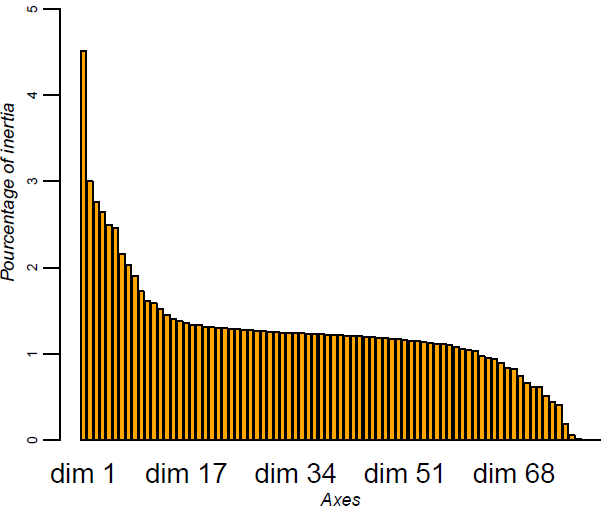}
	\quad
  \includegraphics[width=0.4\textwidth]{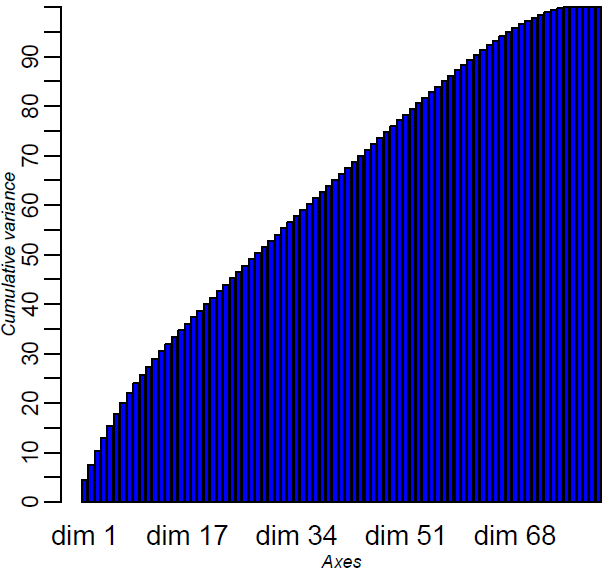}
  \caption{Barplots of the variability (on the left) and the cumulative information captured by the axes (on the right) in the MCA analysis of Adult.}
\label{figAdulteigen}
\end{figure}

Figure~\ref{figAdulteigen} shows the distribution of the variability captured by the axes along with the cumulative level on information. 

\begin{figure}[htbp!]
\centering
  \includegraphics[width=0.45\textwidth]{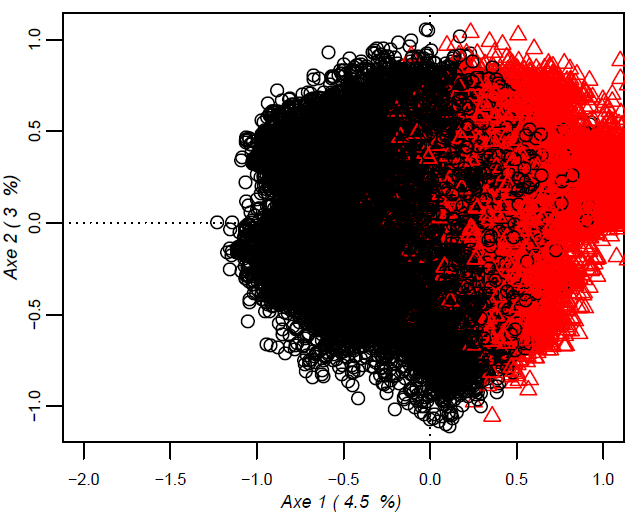}
  \includegraphics[width=0.45\textwidth]{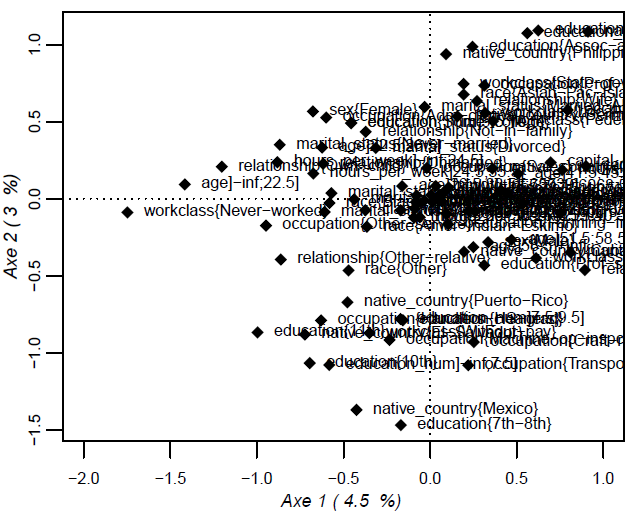}
  \caption{Projection of the set of instances and variable parts, of the Adult database, on the first factorial plan}
\label{figAdultMCA}
\end{figure}

On the contrary to the smaller Iris database, the distribution of the variance (Figure~\ref{figAdulteigen}) indicates that the first two principal axes only capture a cumulative variance of 7.5\%. 
Figure~\ref{figAdultMCA} shows the projections of the instances and variable parts on the first factorial plan  where in the left side figure, the  black circles are the instances that gain less than 50K and the red triangles are the instances that gain more than 50K. Without the  prior knowledge about the class of each instance, which is the case in exploratory analysis,  the projection of instances appears as a single dense cluster. 

The projection on the first factorial plan does not allow to distinguish any clusters, which is not surprising given the low level of variability captured by this plan. However,  in order to capture 20\%, 25\% or 30\% of  the variance, one needs to choose 7, 10 or 13 axes, respectively. Choosing a high number of axes, say 13, means that some post analysis of the projections is required.  

\begin{figure}[htbp]
\centering
  \includegraphics[width=0.45\textwidth]{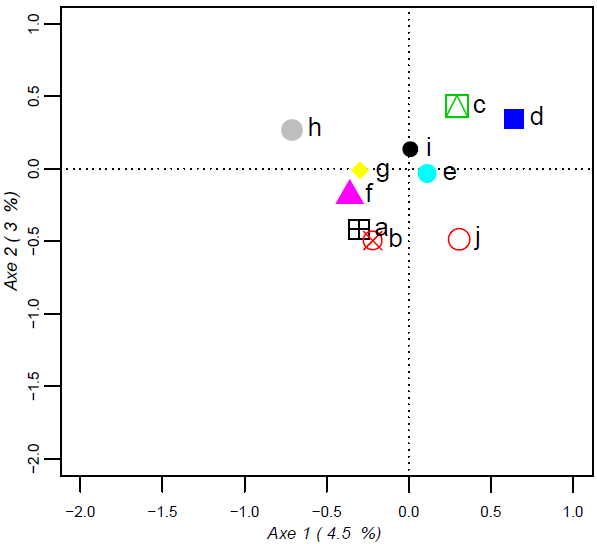}
    \includegraphics[width=0.45\textwidth]{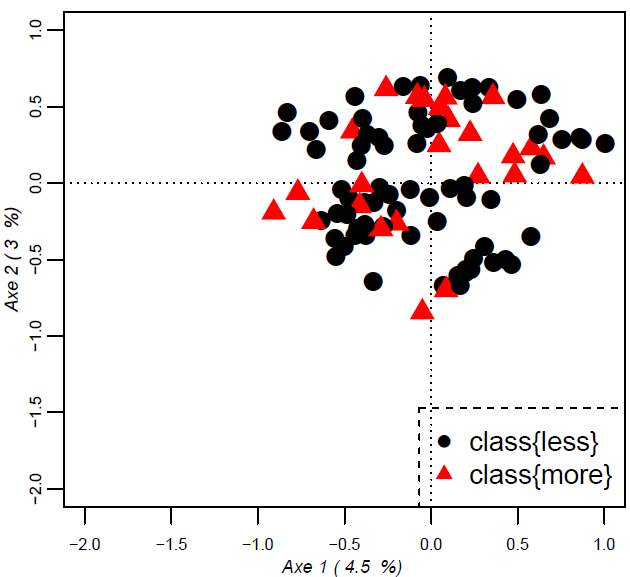}
  \caption{Projection of the k-means centers with k=10 and k=100 clusters, on the first factorial plan.}
\label{project_centers}
\end{figure}

\paragraph{{\it K-means of the MCA projections}}
In order to extract potentially meaningful cluster from the MCA results, we performed a k-means on the projections of the instances and the variable parts on the factor space formed by the first 13 axes. Figure~\ref{project_centers} shows the projection of the k-means centers with $k=10$ (on the left) and $k=100$ (on the right to illustrates how complex the data is).

The k-means clustering of the projections with $k=2$ gives two clusters containing $26178$ instances associated to  50 variable parts, and $22664$ instances associated with 46 variable parts, respectively.  The first cluster of instances associates the variable part \emph{class\{more\}} with being married, white, a men, having   more than 10.5 years of education, being more than 30.5 years old,  working more than 40.5 hours per week, or originating from Canada, Cuba, India  or Philippines. The second cluster of instances associates the variable part \emph{class\{less\}} with being young (\emph{age$]-\infty; 30.5]$}), having less than 10.5 years of education, being never married, divorced or widowed, being Amer-Indian-Eskimo,  black or an other non white race (\emph{race\{Other\}}), working for less than 40.5 hours per week, being a women or originated from countries like El-Salvador, England,  Germany, Mexico, Puerto-Rico, and  United-States. These clusters are consistent with the two main clusters found by the co-clustering, particularly in combining being a men, married, middle aged and working extra hours with earning more than 50K and associating being a women, never married, divorced, or having a child with earning less than 50k.

Table~\ref{summary_adult} shows a summary of the k-means clustering with
$k=10$ indicating the contribution of each cluster to the intra-cluster
variance. To avoid confusion with the clusters resulting from co-clustering,
we name the k-means clusters using letters: $\{a,b, c, d, e, f, g, h, i, j\}$. 

\begin{table}[!htb]
\centering
\begin{tabular}{|l|c|c|c|c|c|c|c|c|c|c|}
\hline
cluster &  $a$ & $b$ & $c$  & $d$ & $e$ & $f$ & $g$ & $h$ & $i$ & $j$\\
\hline
size & 4297 & 1572 & 9325 & 4033 & 2061 & 7686 & 1581 & 4075 & 7163 & 7049\\
\hline
withinss & 4484.7 & 1849.6 & 8185.9 & 3919.8 & 1738.8 & 5156.8 & 1720.7  & 2490.5 & 5447.2  & 3701.6\\
\hline
withinss\% & 11.58  & 4.77 & 21.15 & 10.12  & 4.49 & 13.32 & 4.44  & 6.43 & 14.07 &  9.56\\
\hline
\end{tabular}\caption{Summary of the clusters of instances using k-means}\label{summary_adult}
\end{table}

\begin{table}
\centering
\begin{tabular}{|l|c|c|c|c|c|c|c|c|c|c|} 
\hline         
 cluster &  $a$ &   $b$   & $c$   & $d$  &  $e$  &  $f$   & $g$  &  $h$  &  $i$ &  $j$\\
 \hline
1 & 1679 & 444 & 141& {\bf 2289} & 886  & 12   & 7   & 0  & 48 & 111\\
 \hline
2  &   0  & 20 & 4096 &   0 &   0   & 0 &   {\bf 0}   & 0  &  0   & 0\\
 \hline
3  &   0 &  96  &  0  &  0   & 0 &  18 &   5   & 0  &  0 & {\bf 6377}\\
 \hline
4  &   0 &  31 &  0  &  0 &   0  &  0  &  0  & 13& {\bf 3588} &   0\\
 \hline
5  & 114  & {\bf 88} & 576 & 247 & 129 & 331 &  54 &   28 & 455 & 434\\
 \hline
6  &  0  & 59 &   0 &  0   & 0  &  0 & 252 & {\bf 3314} & 3072  &  0\\
 \hline
7   &  1 & 183 &   0  &  1  &  0 & {\bf 7318} & 776 & 609 &   0 & 127\\
 \hline
8  &   0  & 27 & {\bf 4512}  &  0 &   0  &  3 & 150  & 93  &  0  &  0\\
 \hline
9 & {\bf 2503} & 617   & 0  &  0  &  0   & 2  & 299  & 16  &  0  &  0\\
 \hline
10   & 0 &   7  &  0 & 1496  & {\bf 1046} &   2  & 38 &   2  &  0 &   0\\
\hline
\end{tabular}
\caption{The confusion matrix between the co-clustering and k-means partitions.}\label{confusion_matrix}
\end{table}

Table~\ref{confusion_matrix} shows the confusion matrix between the clusters issued from the co-clustering method and the clusters issued from the k-means of projections. 

The problem of comparing the two clusterings can be seen as a maximum weight matching problem in a weighted bipartite graph, also known as the assignment problem.   It consists of finding the one-to-one matching between the nodes that provides a  maximum total weight. This assignment problem can be solved using the Hungarian method \cite{hungarian1955}. Applied on the matrix of mutual information, the Hungarian algorithm results in the following cluster associations: $(1, d)$, $(2, g)$, $(3, j)$, $(4, i)$, $(5, b)$, $(6, h)$, $(7, f)$, $(8, c)$, $(9, a)$, $(10, e)$ as highlighted in table~\ref{confusion_matrix}. These same associations are also obtained when applying the algorithm to the chi2 table.  This one-to-one matching carries $76.3\%$ of the total mutual information. The highest contributions to the conserved mutual information associate  the k-means cluster $a$  with the co-clustering cluster $9$, the k-means cluster $c$ with the co-clustering cluster $8$, the k-means cluster $f$ with the co-clustering cluster $7$, the k-means cluster $h$ with the co-clustering cluster $6$, the k-means cluster $i$ with the co-clustering cluster $4$, the k-means cluster $j$ with the co-clustering cluster $3$.
 In terms of variable parts, these clusters are as follows: 
\begin{itemize}
\item  the cluster $a$ contains individuals who never-worked or work as handlers-cleaners, have less than 7.5 years of education, or have a level of education from the 7th to the 11th grade. 

\item  the cluster $c$  contains   instances  characterized by:  
    \emph{workclass\{Self-emp-inc\}}, \emph{education\{Assoc-acdm, Bachelors\}}, \emph{education\_num$]11.5;13.5]$}, \emph{occupation \{Exec-managerial, Sales\}}, 
    \emph{race\{Asian-Pac-Islander\}}, \emph{capital\_loss $]77.5;+\infty [$}, \emph{hours\_per\_week$]40.5;48.5]$}, \emph{hours\_per\_week$]48.5;55.5]$}, \emph{native-country\{Germany, Philippines\}}.  

\item  the cluster $f$ contains instances characterized by: earning less than 50K (\emph{class\{less\}}), being relatively young (age$]26.5;33.5]$), having relatively low level of education (\emph{education\{HS-grad\}} and \emph{education\_num$]7.5;9.5]$}), being unmarried, divorced or separated, being an Amer-Indian-Eskimo,  Black or Female. 
			   
\item  the cluster $h$ contains  instances that work less than 35.5 hours per week, are under 26.5 years old, never married and have a child. 

\item  the cluster $i$ contains  middle-aged individuals (between 41.5 and 45.5 years old), with moderate education (9.5 to 10.5 years of education) and  working in farming or fishing. 

\item the cluster $j$ contains instances characterized by the variable parts:  \emph{age$]33.5;37.5]$, age$]37.5;41.5]$, age$]45.5;51.5]$, age$]51.5;58.5]$},
 \emph{workclass \{Self-emp-not-inc\}}, 
 \emph{fnlwgt$]65739;178144.5]$},   
 \emph{hours\_per\_week$]55.5;+\infty [$},  \emph{relationship\{Husband\}},
\emph{marital\_status \{Married-AF-spouse\}}, \emph{marital\_status \{Married-civ-spouse\}}, 
 \emph{occupation\{Craft-repair,  Transport-moving\}},  
 \emph{race-\{White\}}, 
 \emph{sex\{Male\}}.
\end{itemize}
To summarize, the clusters obtained using a k-means on the projections of the
MCA, are somewhat consistent with those obtained using the co-clustering. However, the
process of extracting these clusters, through MCA analysis,  is rather tedious
while with co-clustering, the clusters could be extracted and  explained
simply by looking at the matrix of co-clusters.

\subsection{Discussion}
An important contribution of our methodology, compared to MCA, is its ease of
application and the direct interpretability of its results. When MCA is
applied to a database of a significant size, such as Adult, the projections of
instances and variables on the first factorial plan (and even on the second
plan) do not enable us to distinguish any particularly dense clusters. Therefore,
it is necessary to choose a high number of axes in order to capture enough
information. On the database Adult, we found that 13 axes explain only 30\% of
the information. Choosing this high number of axes meant that some post
analysis of the projections (such as k-means) is necessary to extract any
possible clusters.  Applying this long process for cluster extraction, the
results obtained using k-means, although only explaining 30\% of the
information, are somewhat consistent with those obtained using the
co-clustering and our two-step methodology. However, with our methodology, the
hierarchy of clusters enables us to choose the desired level of detail and the
percentage of information, then one can distinguish, and eventually explain,
the most informative clusters, recognized by their contribution to the total
information.

\section{Conclusion}
\label{conclusion}
In this article, we have proposed a methodology for using co-clustering in
exploratory analysis of mixed-type data. Given a number of parts, chosen by
the analyst, the numerical variables are discretized into intervals with equal
frequencies and the most frequent values in the categorical variables are
kept. A co-clustering between the instances and the binarized variables is
then performed while letting the algorithm infer, automatically, the size of
the summarizing matrix.  

We have shown that on a small database, exploratory analysis reveals a good
agreement between MCA and co-clustering, despite the differences between the
models and the methodologies. We have also shown that exploratory analysis is
feasible even on large and complex databases. The proposed methods is a steps
toward understanding a data set via a joint analysis of the clusters of
instances and the clusters of variable parts.  The results of these
experiments are particularly promising and show the usefulness of the proposed
methodology for real situations of exploratory analysis.

However, this methodology is limited by the need for the analyst to chose a
parameter, the number of parts per variable, used for data
binarization. Furthermore, the co-clustering method does not follow the
origins of the parts which would be useful to consider the intrinsic
correlation structure that exist between the parts originated from the same
variable, which form a partition.  In future work, we will handle these
limitations by defining co-clustering models that integrate the granularity
parameter and track the clusters of variable parts that form a partition of
the same variable. By defining an evaluation criterion for such co-clustering
as well as dedicated algorithms, we hope to automate the choice of the
granularity and improve the quality of the co-clustering results.

\bibliographystyle{apalike}
\bibliography{biblio}

\end{document}